# Areas of Improvement for Autonomous Vehicles: A Machine Learning Analysis of Disengagement Reports

Tyler Ward
Department of Engineering Sciences
Morehead State University
Morehead, United States
tbward@moreheadstate.edu

*Abstract*—Since 2014, the California Department of Motor Vehicles (CDMV) has compiled information from manufacturers of autonomous vehicles (AVs) regarding factors that lead to the disengagement from autonomous driving mode in these vehicles. These disengagement reports (DRs) contain information detailing whether the AV disengaged from autonomous mode due to technology failure, manual override, or other factors during driving tests. This paper presents a machine learning (ML) based analysis of the information from the 2023 DRs. We use a natural language processing (NLP) approach to extract important information from the description of a disengagement, and use the k-Means clustering algorithm to group report entries together. The cluster frequency is then analyzed, and each cluster is manually categorized based on the factors leading to disengagement. We discuss findings from previous years' DRs, and provide our own analysis to identify areas of improvement for AVs.

## I. Introduction

In recent years, there has been a marked increase of autonomous vehicles (AVs) being deployed on roads around the world. Recent studies have found that 18.43 million new cars sold in 2024 will have a level of automation built in that will allow drivers to take their hands off the wheel, and it is estimated that by 2030, 95% of all new vehicles on the market will offer a high or full level of automation [1]. With an estimated 33 million AVs expected to be on the road by 2040 [1], it is becoming increasingly important to understand the limitations of AV technology.

Since 2014, the California Department of Motor Vehicles (CDMV) has monitored and released annual disengagement reports (DRs) for AVs. These DRs contain information from various manufacturers of autonomous vehicles about incidents where their vehicles were disengaged from autonomous mode during driving tests, be it because of a technology failure, or in instances where the test driver/operator had to take manual control of the vehicle to ensure safe operation [2]. This paper analyzes the 2023 reports, although insights gained from previous years' reports are detailed in later sections.

The analysis presented in this paper moves beyond traditional data analysis, instead employing advanced machine learning (ML) and natural language processing (NLP) techniques to gain a deeper understanding of the DR data. Our methodology, which will be detailed in the next section, includes preprocessing, topic modeling, clustering, and categorization techniques. With this paper, we hope to provide a platform for further research into modern limitations of AVs.

## II. Methodology

### A. Combining Datsets

The CDMV provides three different categories of DRs. The first category contains information from DRs where the AV had a driver present in the vehicle, and the vehicle was not capable of operating without a driver, the second category contains information from first-time filers of DRs for AVs, and the third category contains information from DR where the AV was operating fully autonomously, with no driver present.

Each of these categories were provided as separate datasets, so the first stage of preprocessing was to combine these individual datasets into one comprehensive file. This was a simple task, as the features were the same across all three original datasets. Once combined, the conjoined dataset contained 6,584 DRs from 21 AV manufacturers: aiMotive, Apollo, Apple Inc., Aurora Innovation, Bosch, Didi Research America, Gatik, Ghost Autonomy, Imagry, Motional, Nissan USA, Nuro, Pony.ai, Qualcomm, Valeo, Veuron, Waymo, WeRide, Woven by Toyota, Inc., and Zoox. The dataset consists of nine features: 'Manufacturer', 'Permit Number', 'DATE', 'VIN NUMBER', 'VEHICLE IS CAPABLE OF OPERATING WITHOUT A DRIVER (Yes or No)', 'DRIVER PRESENT (Yes or No)', 'DISENGAGEMENT INITIATED BY (AV System, Test Driver, Remote Operator, or Passenger)', 'DISENGAGEMENT LOCATION (Interstate, Freeway, Highway, Rural Road, Street, or Parking Facility)', and 'DESCRIPTION OF FACTS CAUSING DISENGAGEMENT'.

## B. Preprocessing

Given that the DR data was given to the CDMV by each manufacturer themselves, upon manual review it was found that each manufacturer generally followed unique templates for writing descriptions about the factors leading to the disengagement from autonomous mode. This meant that the descriptions from each manufacturer were distinct from each other, but descriptions from the same manufacturer could be repetitive. To make this data easier for our machine learning (ML) models to deal with, the data underwent a preprocessing stage.

Data preprocessing is a set of techniques used prior to the application of a ML model to make the data more suitable for the requirements of the model [3]. To simplify the analysis of the DRs, a short Python script was used to extract only the unique descriptions of events leading to a disengagement from the original dataset. Following this extraction, we were left with 312 unique descriptions from the 6,584 DRs. These 312 descriptions were then used as the input for our ML models.

## C. Topic Modeling

Natural language processing (NLP) is a subfield within ML that deals with the analysis of human language through computational means [4]. Within the context of NLP, topic modeling is a technique used to extract latent variables from a dataset so an NLP system can better understand what events or concepts a document is discussing [5]. Given that the goal for this research is to employ ML for the purpose of analyzing DRs from AVs, topic modeling is an important component in our analysis system, as it facilitates a deeper understanding of the abstract "topics" that appear in each description of disengagement factors in the DRs.

Latent Dirichlet allocation (LDA) is one of the most popular topic modeling methods [5], and it is the one used in this research. LDA assumes that documents are mixtures of topics and that each topic is a mixture of words [5]. This method allows for each document to be described by a distribution of topics and each topic to be described by a distribution of words [5].

To apply LDA in the context of this research, a variety of Python libraries for data manipulation were used: Pandas [6, 7], NumPy [8], NLTK [9], and Gensim [10]. The first step towards implementing LDA was to preprocess the Pandas DataFrame containing the 312 unique descriptions by using NLTK and Gensim to tokenize the text by splitting it into words, convert it to lowercase, remove stop words like "the" and "is", and filter out non-alphabetic words. An example of a description before and after undergoing preprocessing is shown in Table 1.

TABLE I. AN EXAMPLE OF A DESCRIPTION BEFORE AND AFTER PREPROCESSING

| Before Preprocessing | After Preprocessing |
|---|---|
| Safety Driver disengaged autonomous mode upon judging that vehicle was too close to road/lane boundary. Root cause: object, lane detection or other issue. Conditions: Non-inclement weather, dry roads, no other factors involved. | 'safety', 'driver', 'disengaged', 'autonomous', 'mode', 'upon', 'judging', 'vehicle', 'close', 'boundary', 'root', 'cause', 'object', 'lane', 'detection', 'issue', 'conditions', 'weather', 'dry', 'roads', 'factors', 'involved' |

A Python dictionary was then created from the processed texts, mapping each unique word to an integer ID. Tokens that appeared in less than one description or more than half of the descriptions were filtered out, because they were either too rare or too common. Finally, a corpus was created by converting the list of words from each description into a bag-of-words format.

LDA was then applied to the DataFrame. The LDA call was configured to discover ten topics in each description. Once LDA had been performed, a custom function was used to format the topics into a readable format by extracting the dominant topic, its percentage contribution, and the keywords defining the topic for each description. This information was passed into a new DataFrame which was then merged with the DataFrame containing the unique descriptions to create a final DataFrame with the unique descriptions and their associated dominant topic and topic keywords. An example row from this DataFrame is shown in Table 2. Once this DataFrame was created, the topic distribution for each description was calculated and pushed to an array, representing the contribution of each topic to a description.

TABLE II. SAMPLE FROM THE FINAL DATAFRAME

| D[a] | DT[a] | PC[a] | TK[a] |
|---|---|---|---|
| Driver disengaged with steering input. Driver took over because ego vehicle went into a fallback trajectory state immediately after engaging. | 2 | 50.27% | reduce, trajectory, yield, judging, upon, car, state, way, immediately, error |

[a] D = Description, DT = Dominant Topic, PC = Percentage Contribution, TK = Topic Keywords

## D. Clustering

Once the topic distributions were obtained, the next step in the process was to use these distributions as the input to a clustering algorithm. In ML, clustering is a technique to group unlabeled data and extract meaning information from the subsequent clusters [11]. For the purposes of this research, clustering is used to group the unique descriptions together based on the frequency of topics present in them. From there, the clusters will be manually categorized to determine existing challenges in AVs.

This research uses the *k*-Means approach to clustering. The *k*-Means algorithm is a simple and efficient clustering technique that partitions a given dataset into *k* clusters, where each data point belongs to the cluster with the nearest mean [11]. The algorithm seeks to minimize the within-cluster variances, but not the between-cluster variances [11].

One of the most importance considerations for ensuring the best performance of the *k*-Means algorithm is accurately determining the optimal number of clusters, *k*. There are several methods of visually or numerically determining the optimal number of clusters. This research employs the silhouette method for determining the optimal number of clusters.

The silhouette method relies on the use of the silhouette score, which is a measure of how similar an object is to its own cluster compared to other clusters [12]. The silhouette score for each point is a ratio that ranges from -1 to 1 where a value closer to 1 indicates that the point is well inside its cluster and far from neighboring clusters, a value of 0 indicates that the point is on the border of two clusters, and a value close to -1 indicates that the point may have been assigned to the wrong cluster [12]. Fig. 1 shows a scatter plot for the average silhouette scores for values of *k* from 2 to 10.

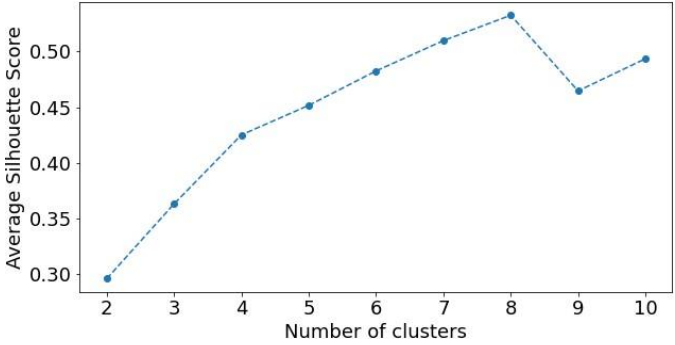

Fig. 1. Average silhouette scores for a range of *k*-values

Because the average silhouette score is highest for a *k*-value of eight, the *k*-Means algorithm is applied to the dataset to group the data points into eight clusters. Once *k*-Means has been applied, the efficacy of the clustering can be evaluated by visualizing the clusters. Because there are eight dimensions to the data, in order to be visualized by traditional means, the number of dimensions needed to be reduced.

Principal component analysis (PCA) is one of the most common methods of dimensionality reduction [13]. However, PCA is a linear reduction technique, and the way data is distributed in the reduced dimensionality after performing PCA may not be accurate to its structure in higher dimensions [13]. To address this issue, this research employs the t-distributed stochastic neighbor embedding (t-SNE) dimensionality reduction technique.

The first stage of t-SNE is the calculation of the similarity between pairs of instances in the high-dimensional space [14]. From there, a similar probability distribution is defined in the low-dimensional space and the Kullback-Leibler divergence between the two distributions with respect to the location of the points in the map is minimized using gradient descent [14]. The visualization of the clusters after undergoing t-SNE dimensionality reduction is shown in Fig 2.

Once the eight clusters were defined, a new feature was added to the DataFrame containing the unique descriptions, mapping each description to their respective clusters. Once the feature was added, a heatmap was generated showing the most common words in each cluster. This heatmap is shown in Fig 3.

Following this visualization, the DataFrame containing the unique descriptions and their respective clusters was merged back into the original DataFrame containing the information from the 6,584 DRs. This resulted in each of the 6,584 data points having an associated cluster, which could then be used to manually categorize the clusters based on the most common words in each cluster to determine areas for improvement in AVs. The frequency of the clusters in the merged DataFrame is represented in the bar chart in Fig. 4.

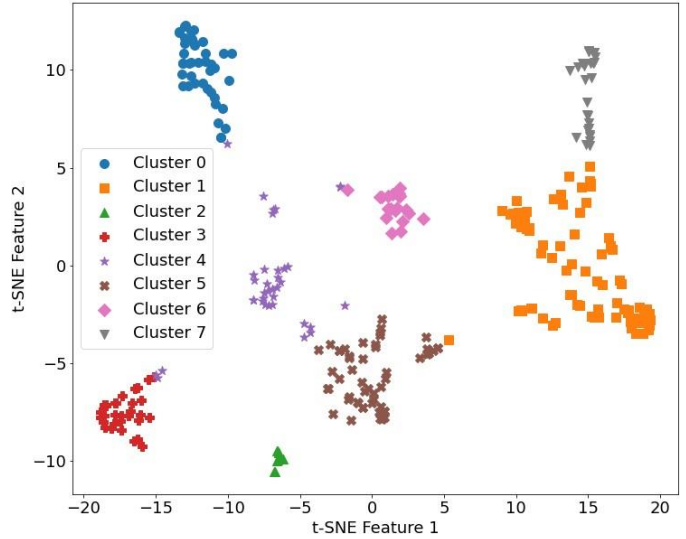

Fig. 2. Visualization of the data clusters after t-SNE

| Words | 0 | 1 | 2 | 3 | 4 | 5 | 6 | 7 |
|---|---|---|---|---|---|---|---|---|
| vehicle | 12 | 112 | 11 | 9 | 16 | 31 | 10 | 40 |
| driver | 30 | 97 | 12 | 2 | 3 | 40 | 15 | 28 |
| took | 0 | 85 | 10 | 0 | 0 | 5 | 3 | 25 |
| velocity | 0 | 76 | 2 | 0 | 0 | 0 | 0 | 25 |
| pedal | 0 | 76 | 0 | 0 | 0 | 0 | 0 | 25 |
| pressed | 0 | 76 | 0 | 0 | 0 | 0 | 0 | 25 |
| increase | 0 | 41 | 0 | 0 | 0 | 0 | 0 | 24 |
| accelerator | 0 | 41 | 0 | 0 | 0 | 0 | 0 | 24 |
| reduce | 1 | 35 | 0 | 0 | 0 | 0 | 0 | 1 |
| brake | 0 | 35 | 0 | 0 | 0 | 0 | 0 | 1 |
| safety | 31 | 0 | 0 | 0 | 3 | 22 | 19 | 0 |
| disengaged | 31 | 11 | 1 | 2 | 6 | 18 | 6 | 1 |
| system | 3 | 0 | 18 | 31 | 3 | 10 | 1 | 0 |
| issue | 28 | 0 | 1 | 0 | 3 | 4 | 0 | 0 |
| due | 25 | 22 | 5 | 2 | 6 | 18 | 8 | 7 |
| braking | 6 | 25 | 2 | 0 | 1 | 10 | 0 | 4 |
| module | 1 | 0 | 0 | 25 | 5 | 2 | 0 | 0 |
| planning | 3 | 1 | 0 | 5 | 24 | 9 | 1 | 1 |
| software | 1 | 0 | 3 | 24 | 4 | 0 | 0 | 0 |
| lane | 10 | 22 | 9 | 14 | 5 | 5 | 4 | 23 |
| av | 2 | 0 | 19 | 9 | 2 | 23 | 2 | 0 |
| turn | 3 | 22 | 0 | 1 | 1 | 1 | 0 | 0 |
| weather | 21 | 0 | 0 | 21 | 2 | 0 | 5 | 0 |
| failure | 0 | 7 | 0 | 21 | 1 | 0 | 0 | 1 |
| perception | 9 | 0 | 3 | 19 | 4 | 9 | 0 | 0 |
| change | 4 | 6 | 0 | 2 | 2 | 1 | 0 | 19 |
| control | 1 | 0 | 13 | 0 | 0 | 19 | 0 | 0 |
| logic | 0 | 0 | 0 | 0 | 19 | 3 | 0 | 0 |
| left | 0 | 18 | 0 | 1 | 1 | 2 | 0 | 0 |
| test | 11 | 0 | 17 | 5 | 0 | 2 | 1 | 0 |

Fig. 3. Heatmap showing the most common words per cluster

### E. Categorization

With the clusters merged back with the original DataFrame, the next phase of the analysis phase was to manually review the information obtained from the clusters, and identify notable challenges in AVs based on the categorization of the clusters by the most common words found in each cluster. The categories these clusters were grouped into are defined below:

- *Cluster 0 – Perception and Timing Failures*: Primarily focused on safety disengagements due to issues with perception, inappropriate timing, and braking.

- *Cluster 1 – Complex Navigation Difficulties*: Spans a broad range of navigation-related issues, such as turning, lane changes, and speed adjustments.

- *Cluster 2 – Sensor and Tracking Malfunctions*: Deals with technical malfunctions or limitations, specifically in terms of sensor placement, unexpected sensor readings, and loss of tracking.

- *Cluster 3 – Adverse Condition System Failures*: Relates to software or system failures, particularly under specific conditions like adverse weather.

- *Cluster 4 – Multifactorial Incident Spectrum*: This cluster is a mix of several topics, indicating incidents in complex scenarios involving multiple factors, such as software failures, safety disengagements, and perception issues.

- *Cluster 5 – Safety Protocol Deviations*: Focuses on discrepancies between autonomous mode decisions and expected safe behaviors.

- *Cluster 6 – Varied Navigation and Control Issues*: Covers a range of issues involving navigation and control, such as unexpected behaviors relating to ghost braking.

- *Cluster 7 – Specific Navigation Challenges*: Focuses on specific scenarios or types of maneuvering difficulties, such as the cause of navigation errors, the types of maneuvers involved, or the conditions under which these issues arise.

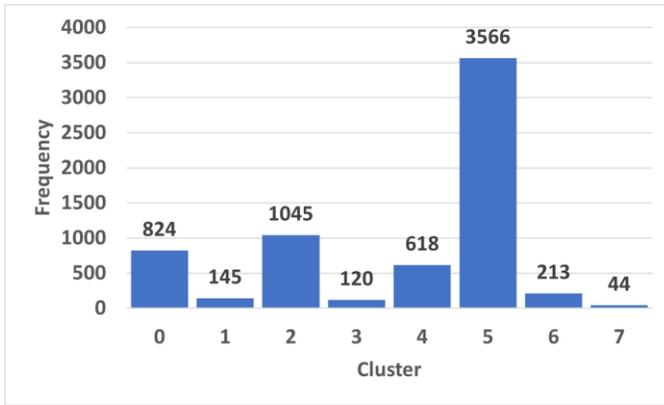

Fig. 4. Frequency of clusters in the final DataFrame

With these categories identified, the analysis can begin. The next section of this paper details existing research in this area, drawing from insights from previous years AV DRs released by the CDMV. Following this, there will be a discussion of the information obtained from the clusters, and conclusions regarding the state of modern AV systems will be drawn.

## III. PAST INSIGHTS

The first dataset of DRs from AVs released by the CDMV contained information from September 2014 to November 2015. The authors of [15] analyzed this data in an attempt to gain insight into the factors influencing disengagement from autonomous mode. They identified a strong correlation between autonomous miles driven and accidents, highlighting the potential of autonomous miles as a risk assessment measure for disengagements and accidents. The study also underscored the impact of trust on driver engagement, with a lack of trust leading to quicker manual intervention. The authors highlighted the importance of further research into human-machine interfaces, driver expectations, and the psychological aspects of AV operation.

A subsequent study [16], analyzed the same set of data as the previous, but also included the next year's data, up to January 2017. A notable finding from this study was that disengagements rarely lead to accidents, with an average of one accident per 178 disengagements. This study highlighted the importance of analyzing disengagements as potential indicators of future accidents, while criticizing the regulatory framework for AV testing in California, pointing out limitations in the regulation's wording and structure that hinder the clarity and usefulness of the reported data.

Another study analyzed the DRs up to November 2018 [17]. Their findings showed that AV systems are less likely to disengage on streets and roads compared to freeways and interstates, suggesting that complex urban environments with diverse interactions pose fewer unforeseen challenges to the AVs than high-speed, less complex freeway environments. The authors also found that disengagements are more likely due to hardware and software discrepancies, planning errors, or environmental factors and interactions with other road users, highlighting the limitations of AV systems in processing and responding to real-world scenarios compared to human drivers.

DR data up to 2019 was analyzed in [18]. Notably, a key finding of this survey was that automated disengagement events tend to decrease with the accumulation of experience and miles driven autonomously. This is notable because it is opposition with the findings of [15], indicating that over time, the accumulation of autonomous miles driven plays less of a factor in disengagements as technology has improved. However, while the number of autonomous disengagements has decreased, the authors of [18] found that the rate of manual disengagement has remained high, indicating the continued lack of human trust in AV technology, which is in-line with the findings of [15].

The findings of [15, 18] were further verified by [19], which examined the DR data up to 2020. The authors found that 80% of the disengagements were initiated by test drivers, who either felt uncomfortable about the maneuver of the AVs or made precautionary takeovers because of insufficient trust. This study also suggested that discrepancies in perception, localization, mapping, planning, and control were the primary causes that led to the AV struggling to perform certain tasks.

An analysis of the DR data from 2017 to 2021 [20] found that factors related to human error, system failure, surrounding vehicles, and roadway failures could cause an AV-involved pre-crash disengagement. One study sought to use the DR data up to 2022 to create virtual test scenarios for improving the performance of AVs under certain conditions [21]. However, they found that the DR data was very repetitive and, in many cases, did not contain enough information to be able to reconstruct the situation causing the disengagement.

## IV. ANALYSIS

Now that information has been gathered from the 2023 AV DR data and past insights have been discussed, the question remains: What are the current limitations and challenges towards the use of AVs on public roads? From the cluster analysis, it is apparent that cluster five is by far the most represented cluster, with over half of the DRs belonging to this cluster. As mentioned, this cluster was categorized as describing incidents where the decisions made by the AV in autonomous mode deviated from expected safe behaviors. However, this description is fairly broad, and needs to be analyzed further to truly be representative of current AV challenges and limitations.

Using the methods described previously, this cluster was isolated for further analysis. It was found that the cause of disengagements in this cluster could be categorized into more specific categories: hardware issues, motion planning and control issues, incorrect predictions, perception issues, localization issues, incorrect maps, issues in the recording module, incorrect router transitions, planning issues, deviance from expected behavior, and a hybrid category containing descriptions with multiple events that led to a disengagement. A pie chart showing a breakdown of issues found in combination in the hybrid category is shown in Fig. 5.

From the pie chart, we can see that six of the eight combinations of factors from the hybrid category include issues with the motion plan of the AVs. This indicates that one of the major problems with modern AV systems is flawed decision making in terms of the movement of the vehicle after an issue in a dependent component. This signals the need for further research into understanding the interconnectivity of components of an AV, and how an error in one component can affect the operation of another.

Once merged back into the full dataset containing cluster five, the final frequency of the categories is shown in Fig. 6. We can see that of the 3,566 DRs in this cluster, 1,926 of the disengagements were caused by incorrect predictions leading to a dissatisfactory motion plan. This indicates that there is still a long ways to go in terms of being able to produce fully automous cars that can exist simultaneously with human drivers in the road.

## V. CONCLUSION

Cluster analysis of the 2023 AV DRs released by the CDMV revealed distinct categories of challenges faced by modern AVs, ranging from perception and timing failures to complex navigation difficulties, sensor malfunctions, adverse condition system failures, multifactorial incident spectrums, safety protocol deviations, varied navigation and control issues, and specific navigation challenges. The cluster regarding deviations from safety protocols was the most represented cluster, with ~54% of the DRs falling into this cluster. Further analysis of this cluster identified specific causes of disengagements, such as issues related to hardware, motion planning and control, predictions, perception, localization, maps, recording modules, router transitions, planning, and deviance from expected behavior.

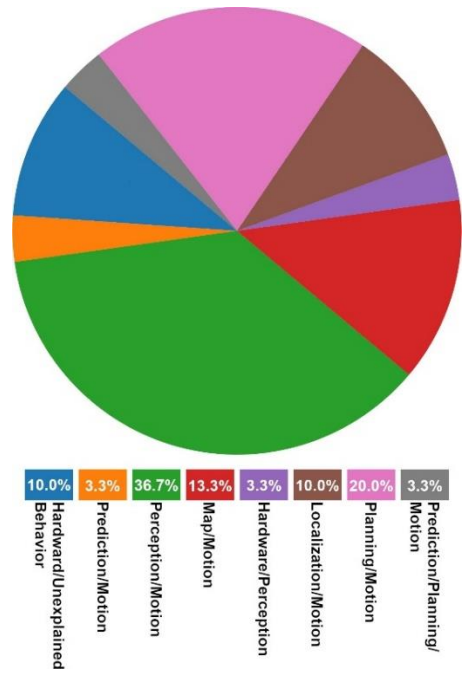

Fig. 5. Statistics on combined factors leading to disengagement from autonomous mode

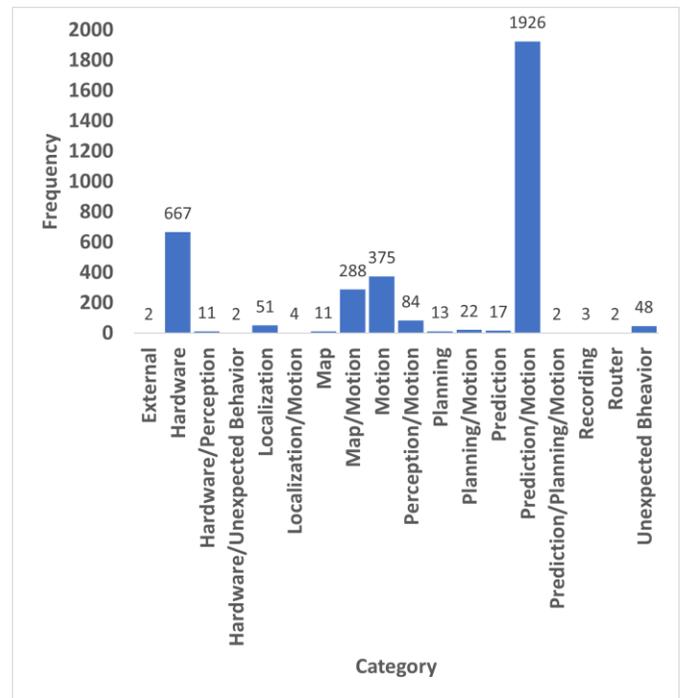

Fig. 6. Final frequency of the categories of disengagement

A notable finding was the prevalence of failures in other components leading to the AV generating unsatisfactory motion plans. Specifically, in circumstances where the AV incorrectly predicted an outcome, the odds of an unsatisfactory motion plan being generated were significantly higher. This observations indicates the need for further research and development aimed at enhancing the decision-making capabilities of AVs to mitigate risks and improve their compatibility with human drivers on public roads. The analysis of the 2023 DR data highlighted some of the existing limitation and challenges of AV development. Addressing these challenges will require a multifaceted approach of technological advancements, a deeper understanding of the interconnectivity within AV systems, the development of rigorous testing protocols, and regulatory frameworks focused on enhancing the safety of these vehicles.